\documentclass[lettersize,journal]{IEEEtran}
\usepackage{amsmath,amsfonts}
\usepackage{algorithmic}
\usepackage{algorithm}
\usepackage{array}
\usepackage[caption=false,font=normalsize,labelfont=sf,textfont=sf]{subfig}
\usepackage{textcomp}
\usepackage{stfloats}
\usepackage{url}
\usepackage{verbatim}
\usepackage{graphicx}
\usepackage{cite}

\usepackage{xcolor}  
\usepackage{graphicx}
\usepackage{float}
\usepackage{cite}
\usepackage{tabu}                     
\usepackage{multirow}                 
\usepackage{multicol}                 
\usepackage{multirow}                
\usepackage{float}                    
\usepackage{makecell}                 
\usepackage{booktabs}                 

\usepackage{soul, color, xcolor}

\begin{document}
	
	\title{RS-DGC: Exploring Neighborhood Statistics for \\ Dynamic Gradient Compression on Remote Sensing Image Interpretation}
	
	\author{Weiying Xie, \IEEEmembership{Senior Member,~IEEE}, Zixuan Wang, Jitao Ma, Daixun Li, and Yunsong Li, \IEEEmembership{Member,~IEEE}
		\thanks{This work was supported in part by the National Natural Science Foundation of China under under Grant 62121001, Grant 62322117, Grant 62371365 and Grant U22B2014; in part by Young Elite Scientist Sponsorship Program by the China Association for Science and Technology under Grant 2020QNRC001. (\textit{Corresponding author: Zixuan Wang}.)}      
		\thanks{W. Xie, Z. Wang, J. Ma, D. Li and Y. Li are with the State Key Laboratory of Integrated Services Networks, Xidian University, Xi'an 710071, China (e-mail: wyxie@xidian.edu.cn; zxwang1002@stu.xidian.edu.cn; 21011210271@stu.xidian.edu.cn; ldx@stu.xidian.edu.cn; ysli@mail.xidian.edu.cn).}
	}
	
	\markboth{IEEE Transactions on Geoscience and Remote Sensing}%
	{Shell \MakeLowercase{\textit{et al.}}: A Sample Article Using IEEEtran.cls for IEEE Journals}
	
	\IEEEpubid{0000--0000/00\$00.00~\copyright~2023 IEEE}
	
	\maketitle
	
	\begin{abstract}
		Distributed deep learning has recently been attracting more attention in remote sensing (RS) applications due to the challenges posed by the increased amount of open data that are produced daily by Earth observation programs. However, the high communication costs of sending model updates among multiple nodes are a significant bottleneck for scalable distributed learning. Gradient sparsification has been validated as an effective gradient compression (GC) technique for reducing communication costs and thus accelerating the training speed. Existing state-of-the-art gradient sparsification methods are mostly based on the “larger-absolute-more-important” criterion, ignoring the importance of small gradients, which is generally observed to affect the performance. Inspired by informative representation of manifold structures from neighborhood information, we propose a simple yet effective dynamic gradient compression scheme leveraging neighborhood statistics indicator for RS image interpretation, termed RS-DGC. We first enhance the interdependence between gradients by introducing the gradient neighborhood to reduce the effect of random noise. The key component of RS-DGC is a Neighborhood Statistical Indicator (NSI), which can quantify the importance of gradients within a specified neighborhood on each node to sparsify the local gradients before gradient transmission in each iteration. Further, a layer-wise dynamic compression scheme is proposed to track the importance changes of each layer in real time. Extensive downstream tasks validate the superiority of our method in terms of intelligent interpretation of RS images. For example,  we achieve an accuracy improvement of 0.51$\%$ with more than 50$\times$ communication compression on the NWPU-RESISC45 dataset using VGG-19 network. To the best of our knowledge, this is the first gradient compression method designed for RS images and downstream tasks, achieving a successful trade-off between high compression ratio and performance.
	\end{abstract}
	
	
	\section{Introduction}
	\IEEEPARstart{W}{ith} the rapid advancements in satellite technology and sensor technology, researchers can now effortlessly gather a vast amount of high-quality remote sensing (RS) images. This makes it a perfect target area for data-driven applications. In recent years, developments in software and hardware technology have greatly influenced Earth observation applications, particularly in RS image interpretation and deduction. These advancements have facilitated access to datasets of superior quality, at faster acquisition rates \cite{haut2021distributed}. These RS data can be used for downstream tasks such as object detection, image classification and land cover classification, and are of great significance in urban planning, agriculture, environmental monitoring and other fields \cite{li2023fedfusion, veraverbeke2018hyperspectral, appel2018open, bioucas2013hyperspectral}. As a method to accelerate deep learning, distributed learning \cite{bertsekas2003parallel} can further help us solve the problem of intelligent interpretation of RS data efficiently, especially in the age of RS big data \cite{zhang2019remotely}. Consequently, significant efforts have been made to develop a multi-training nodes constellation system aimed at obtaining more efficient and comprehensive results in the analysis of RS data.
	\begin{figure}[t]
		\centering
		\includegraphics[width=0.9\columnwidth]{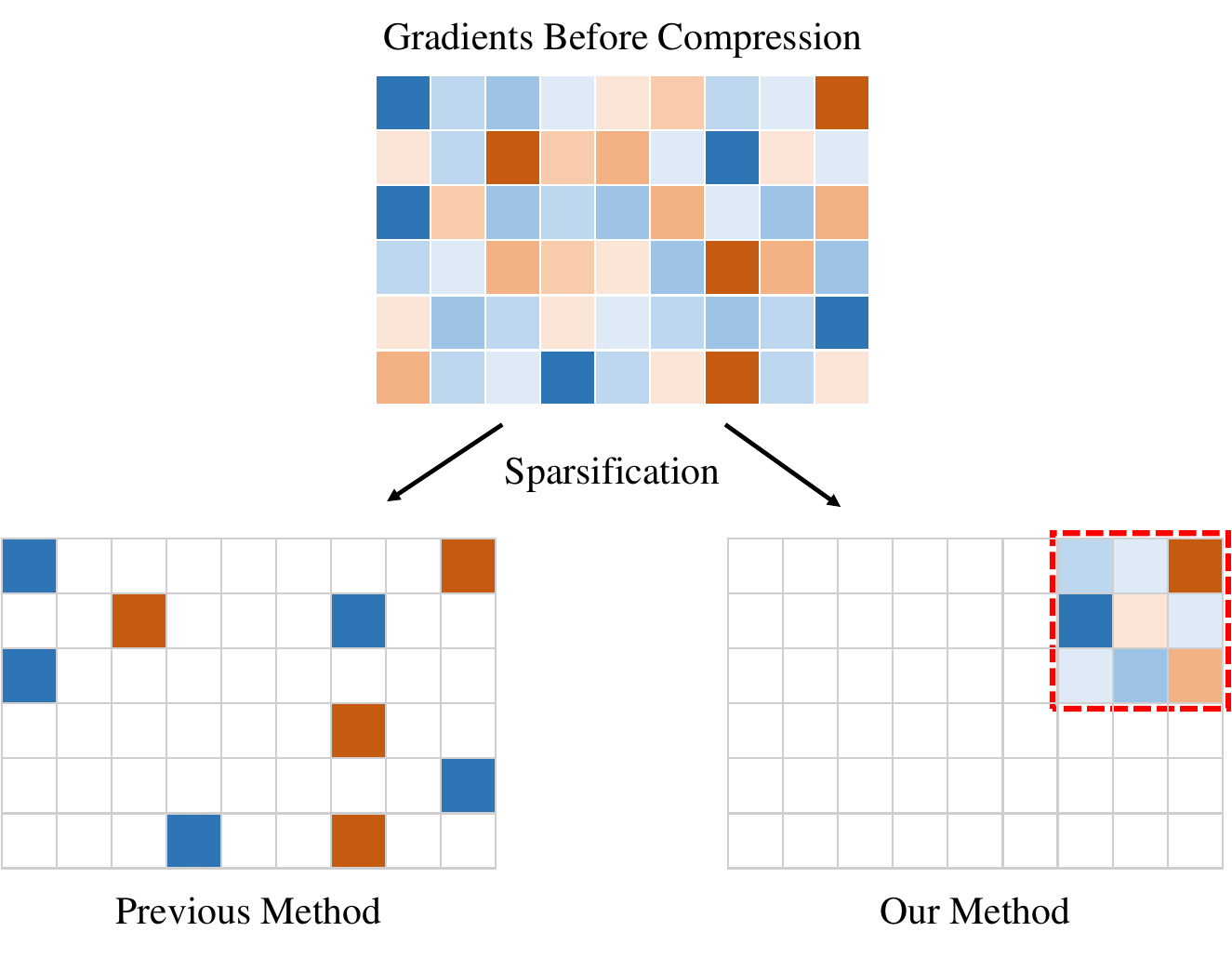} 
		\caption{An illustration of the compressing criterion for absolute-based approach and the proposed method. We interpret a layer's gradients as a matrix. Coordinate values are color coded (\textcolor{orange}{positive}, \textcolor{blue}{negative}), where deeper color denotes larger absolute value of the gradient. For the previous criterion, only the gradients with the largest absolute value are kept. In contrast, RS-DGC selects gradient neighborhoods that contribute more to network updating.}
		\label{fig1}
		\vspace{-0.5cm} 
	\end{figure}
	\IEEEpubidadjcol   

	However, the communication costs keep rising with more and more nodes. This leads to notable delays in transmission, significantly affecting the training speed within a satellite distributed platform that is already constrained by limited resources \cite{review}.

	In recent years, the swift progress of deep neural networks (DNNs) has provided technical support for a wide range of visual tasks in the field of RS \cite{bi2019apdc, zhu2019attention, zhang2019remote, xu2020multilayer}. Its powerful image processing capabilities have been successfully used in complex scene interpretation. However, with the increasing complexity and fineness requirement of target recognition in RS tasks, the scale of deep learning models is also increasing exponentially, leading to a large amount of redundancy in parameter transmission between nodes. More seriously, the high communication costs required for over-parameterization hinders the application of RS technology in scenarios with limited physical space.
	
	Gradient compression (GC) has emerged as an effective approach to reduce communication costs during distributed learning. Many effective techniques have been proposed to compress gradients, such as gradient sparsification \cite{sparse1,sparse2,sparse3,threshold-v}, quantization \cite{qsgd,signsgd,Hyper-Sphere-Quantization,Error-Feedback-Fixes-SignSGD,ecqsgd,Terngrad}, and low-rank decomposition \cite{GradZip,ATOMO,powersgd,GradiVeQ}. Among them, gradient sparsification transmits only a fraction of the gradient elements for full ones without compromising accuracy, thus offering a higher compression ratio (e.g., 0.1$\%$ in DGC \cite{DGC}) compared to gradient quantization and low-rank decomposition. DGC introduced local gradient accumulation and momentum correction, reducing the communication costs in a more aggressive way. Nevertheless, DGC ignored the significance of small gradients, resulting in a certain degree of accuracy loss.
	
	Current practice \cite{CSER,Qsparse-local-SGD} performs GC by following the “larger-absolute-more-important” criterion, which believes that gradients with smaller absolute values can be sparsed safely due to their less importance. As shown in Fig. \ref{fig1}, after calculating the absolute values of gradients in a model, a pre-specified or pre-computed threshold $v$ is utilized to select gradients whose absolute value is smaller than it.
	However, this criterion ignores the effect of small gradients on model updating, leading to some relatively important neurons being unable to be updated for a long time. 
 	From a network training perspective, although large gradients can be effective in model updating, they often lead to challenging stability of convergency that degrade the model’s performance. Furthermore, most existing GC methods treat each gradient as an independent individual and evaluate its importance. Yet, they ignore the influence of neighborhood information. Gaining contextual information from the neighborhood could learn informative structure-aware node representations \cite{neighbor}. The gradient's neighborhood information encompasses nearby gradient direction and magnitude, offering valuable insights for gradient compression and reconstruction. Leveraging this information enhances our comprehension of the gradient's distribution and trends, leading to minimizing distortion in the compressed gradients.  
	
	Despite the possibility of noise or inaccuracies, it's crucial to emphasize that small gradients should neither be disregarded nor replaced. These seemingly diminutive gradients, despite their small magnitudes, harbor essential information vital to the training process. They actively contribute to the convergence of models and significantly influence achieving optimal performance. Previous studies have also pointed out that we should take the exploration of gradient information (i.e., sampling small gradients) into account \cite{bayes,gradient_information1,gradient_information2}. Employing these smaller gradients reasonably can enhance model stability and generalization, providing a safeguard against overfitting and local minimum challenges. 
	
	To address these challenges, we propose RS-DGC, a dynamic gradient compression method by exploring the neighborhood statistics information for RS image interpretation. Some high-dimensional data, especially RS data, is actually a low-dimensional manifold structure embedded in a high-dimensional space \cite{embedding}. Inspired by this, we introduce neighborhood statistics indicator (NSI), a simple yet powerful indicator, to obtain the sparse representation of raw gradients efficiently. Different from the previous methods which sparsed gradients with small absolute, RS-DGC chooses the gradients with the most information. Specifically, we calculate the NSI by computing the mean and standard deviation of gradients within a specified neighborhood. Employing the NSI allows us to quantitatively assess gradient importance, capturing both relatively small but significant gradients and larger ones. This strategy ensures that the gradients that need to be transmitted offer richer information and are more conducive to model updating. 
	Yet, due to the dynamic nature of DNNs training, the importance of gradients across various layers fluctuates throughout the process. A fixed compression ratio fails to dynamically monitor the evolving significance of each layer, resulting in a certain degree of accuracy compromise. Consequently, a real-time dynamic compress policy that can adapt to the dynamic convergence process must be developed.

	We thoroughly evaluated the effectiveness and generalization of our proposed method by performing extensive experiments on two downstream tasks of RS. These tasks include RS image classification and HSI/LiDAR multi-modal scene classification. To assess the performance of our method, we employ various neural networks such as convolutional neural networks (CNNs) and vision transformer (ViT), and conduct comparative experiments against state-of-the-art algorithms. The experimental results indicate that our approach exhibits competitiveness, particularly in scenarios with high sparsity compared to existing gradient compression methods.
	
	The major contributions of our work are outlined as follows:
	
	\begin{itemize}
		\item[$\bullet$]	We propose a new GC criterion called NSI. To our knowledge, it is the first time that taking advantage of gradient neighborhood to quantify the importance of gradients rather than individual gradients. NSI eliminates the contingency of individual gradients to model update by obtaining contextual information and captures the most informative gradient through gradient statistical indicators, which enables a more stable compressing.
		\item[$\bullet$]	 We propose a layer-wise dynamic adaptive compression ratio adjustment strategy to track the importance changes of each layer. It can adaptively adjust the compression ratio of each layer in the process of model training with weight information. Not only does this method help to enhance model accuracy, but the compression ratio can also be maintained at a very high level.
		\item[$\bullet$]	Extensive experiments on multiple neural networks and RS downstream tasks are conducted to show the effectiveness of our proposed RS-DGC. Experimental results show that our method is superior to mainstream techniques in terms of accuracy and communication costs.
	\end{itemize}
	
	The rest of this article is organized as follows. In Section \ref{p2}, we introduce the related works. In Section \ref{p3}, we describe the proposed approach in detail. In Section \ref{p4}, a series of experiments on several downstream tasks are reported.
	Finally, the conclusions of the study are drawn in Section \ref{p5}.

	\section{Related Works}
	\label{p2}
	\subsection{Gradient Compression}
	Broadly speaking, GC methods can be roughly divided into three main categories: gradient quantization, low-rank decomposition and gradient sparsification. We will introduce them respectively.
	\subsubsection{Quantization}
	Quantization methods reduce the number of bits of each element of the gradients by performing a low bit representation of float32 data of the original gradient. SignSGD \cite{signsgd} quantize each float32 element of the gradient to 1-bit sign, by mapping the negative components to -1 and the others to +1. Seide \textit{et al.} \cite{seide20141} proposed an error compensation scheme to reduce the error caused by quantization. Each time, the quantization error of the previous gradient is added to the current gradient, and then quantized. Q-SGD \cite{qsgd} quantize each element via randomized rounding. Wu \textit{et al.} \cite{ecqsgd} proposed ECQ-SGD, which also utilized the error compensation method. Unlike 1-bit SGD, ECQ-SGD incorporated the accumulation of all previous quantization errors as the compensation error. 
	However, quantization can at most reduce the communication costs by 32$\times$. And because the low-bit representation of gradients will lose information, it is almost impossible to avoid accuracy loss.
	
	\subsubsection{Low-rank Decomposition}
	Low-rank decomposition pertains to the process of decomposing the original gradient tensor into multiple matrices or tensors of lower rank. Wang \textit{et al.} \cite{ATOMO} employed computationally intensive singular value decomposition (SVD) to accomplish low-rank decomposition, whereas Vogels \textit{et al.} \cite{powersgd} utilized power iteration to calculate lower-rank matrices with significantly reduced computational costs for decomposition. However, the compression ratio of low-rank decomposition is also very limited.
	
	\subsubsection{Sparsification}
	Gradient sparsification is a highly promising approach for reducing communication overhead compared to the aforementioned methods. By leveraging sparsity, this method effectively lowers communication costs by transmitting only a small subset of gradients. Random-$k$ \cite{random-top-k} sparsifiers randomly selects $k$ elements from the gradients for communication, whereas Top-$k$ \cite{top-k} sparsifiers selects elements with higher magnitudes by sorting the gradients for transmission. Compared to Top-$k$ sparsifiers, Random-$k$ sparsifiers exhibit higher randomness, which could not guarantee the selection of larger gradients. Besides, there are many other ideas for gradient sparsification. Since gradients between different nodes have high mutual information, Abrahamyan \textit{et al.} \cite{LGC} took advantage of inter-node redundancy to improve compression efficiency by training auto-encoder (AE) to compress gradient information and decompress it on other nodes. Tao \textit{et al.} \cite{CE-SGD} proposed CE-SGD, which could adaptively adjust the gradient sparsity according to the model's feedback. It also selectively transmitted the gradients based on their degree of participation in the back propagation.

	\subsection{RS Distributed Learning}
	The explosive growth of RS data presents significant challenges for effectively processing data to support various remote sensing applications \cite{wang2014estimating, rathore2015real, syrris2018mosaicking, yuan2020deep}. With the increasing volume and speed of data generation in RS, processing has inevitably become a big data problem \cite{ma2015remote, chi2016big, sun2019efficient}. As a result, numerous recent studies have concentrated on the distributed parallel processing of RS big data.
	Boulila \textit{et al.} \cite{boulila2021rs} proposed RS-DCNN, a RS large datasets classification method based on distributed deep learning, which performs data parallel training on the Apache Spark framework.
	Martel \textit{et al.} \cite{martel2016gpu} proposed a parallel algorithm called pFUN for full unmixing of remotely sensed hyperspectral images on two GPUs by using CUDA programming and the cuBLAS library, respectively. Certain computation-intensive operations in the unmixing flow are performed on these two GPUs to enhance the computational efficiency of the entire processing flow.
	Sun \textit{et al.} \cite{sun2019efficient} proposed a new big data framework to process large amounts of RS images on cloud computing platforms. In addition to leveraging the parallel processing capabilities of cloud computing to deal with large-scale RS data, this framework also incorporates task scheduling strategies to further exploit the parallelism of the distributed processing stage.

	\begin{figure*}[t]
		\centering
		\includegraphics[width=2.1\columnwidth]{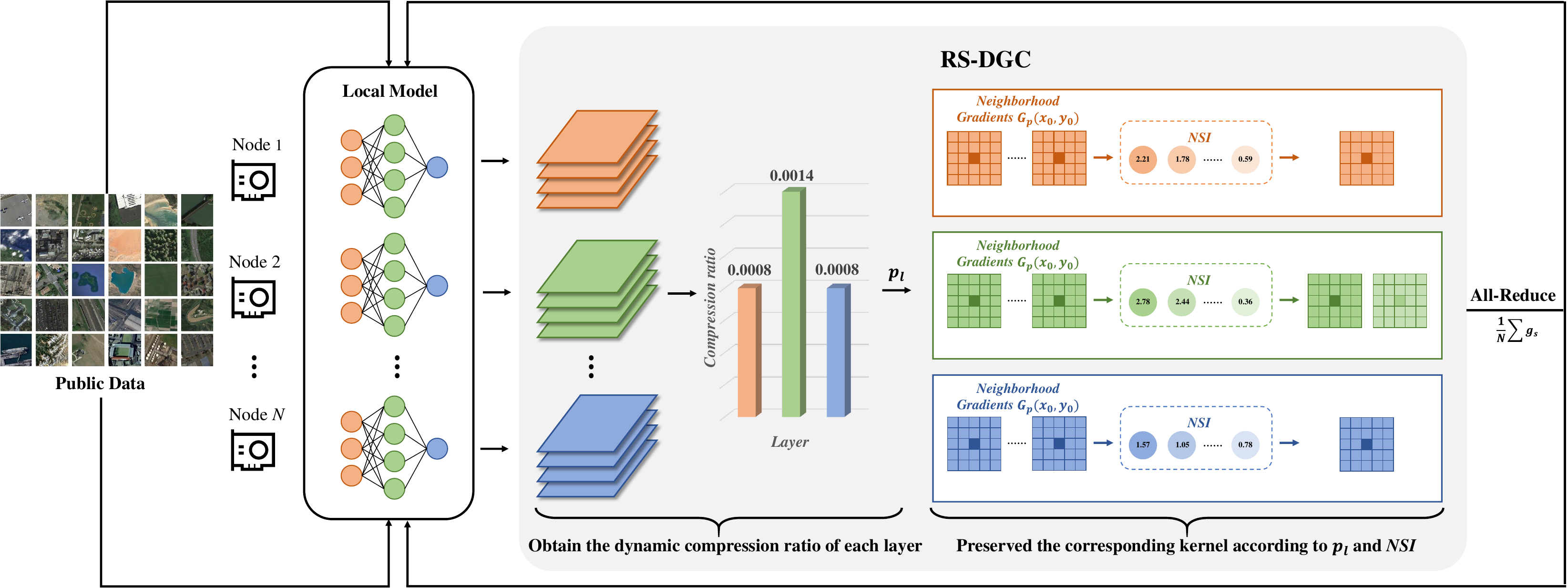} 
		\caption{Main workflow of RS-DGC. Firstly, utilizing the dynamic compression ratio strategy, each node calculates the compression ratio for every layer of the local network. Then, each node undertakes independent local training using its own data. After the backward propagation, gradient compression is performed based on the NSI of each neighborhood and the corresponding compression ratio corresponding to this layer.}
		\label{fig2}
	\end{figure*}
	
	\section{Methodology}
	\label{p3}
	This section describes the formulation of our GC method. First, we introduce the formulation of our problem. Then, we describe the dynamic compression strategy. Finally, we introduce the gradient compression method based on NSI. 
	
	\subsection{Problem Definition}
	A distributed learning system consists of $N$ nodes and each node can perform local training based on RS dataset $D_n$ ($1 \leq n \leq N$). The purpose of GC is to reduce the gradient elements that need to be transmitted as much as possible without losing the accuracy of the model, thereby minimizing the communication costs. Our aim is to solve the following optimization problem:
	\begin{equation}\label{eq1}
		\begin{aligned}
			\min & \sum_{n=1}^N \mathcal{C}\left(\Phi\left(g_n, p\right)\right) \\
			\text {s.t.} & \frac{1}{N} \sum_{n=1}^N\mathcal{L}_{\mathrm{SGD}}\left(g_n, D_n\right) \geq \frac{1}{N} \sum_{n=1}^N \mathcal{L}_{\mathrm{S}-\mathrm{SGD}}\left(\Phi\left(g_n, p\right), D_n\right)
		\end{aligned},
	\end{equation}
	where $\mathcal{C(\cdot)}$ represents communication cost, $\Phi(\cdot)$ represents the operation to the gradients of each node. $\mathcal{L}_{\mathrm{SGD}}$ is the loss function using stochastic gradient descent (SGD) and $\mathcal{L}_{\mathrm{S}-\mathrm{SGD}}$ is the loss function after gradient compression. $g_n$ is gradients at node $n$, $p$ is the global compression ratio.
	
	The gradient sparsification and synchronization using classic data-parallel SGD can be formulated as:
	
	\begin{equation}\label{eq2}
		\begin{aligned}
			g_t & =\nabla f_n\left(\mathrm{X}_t ; D_n\right), \quad g_t^s=\operatorname{Sparse}\left(g_t\right), \\
			g_t^* & =\operatorname{Sync}\left(g_t^s\right), \quad \mathrm{X}_{t+1}=\mathrm{X}_t-\eta g_t^*.
		\end{aligned}
	\end{equation}
	
	At each iteration $t$, $\mathrm{X}_t$ is the model parameters, $g_t$ is local raw gradients, $g_t^s$ is the local gradients after sparsification and $g_t^*$ is global gradients after synchronization. $\eta$ represents the learning-rate.
	
	\subsection{Dynamic Compression Strategy}
	When considering DNNs, it is crucial to acknowledge that the significance of each layer within the model is not uniform. We posit that layers with larger weight parameters hold greater importance. This proposition is primarily rooted in the fact that weight parameters dictate the extent to which each layer processes input data. By employing larger weight values, the input data can exert a more pronounced influence on subsequent layers, thereby amplifying the importance of the given layer. Conversely, smaller weight values attenuate the impact of input data, thereby diminishing the layer's contribution to the overall network. Furthermore, the utilization of larger weight values may result in the propagation of more substantial gradient values throughout that layer during the backward propagation. Consequently, it becomes necessary to determine distinct compression rates for individual layers. Additionally, the importance of each layer can experience variations during the neural network training process. 
   Consequently, we derive real-time dynamic compression rates for each layer, which are subsequently employed in subsequent sparsification procedures.

\begin{figure*}[t]
		\centering
		\includegraphics[width=2\columnwidth]{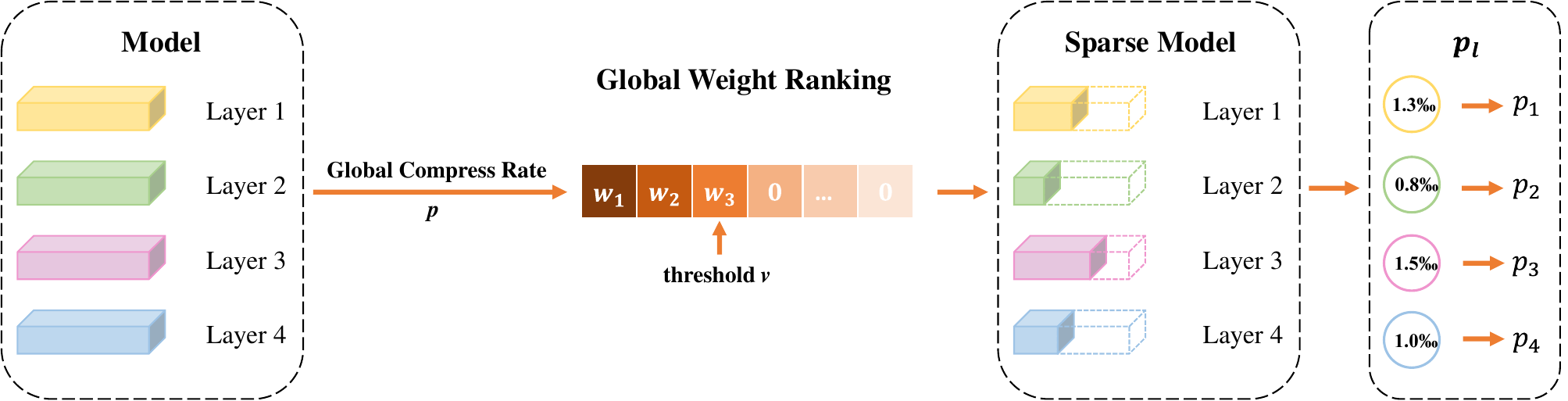} 
		\caption{Framework of dynamic gradient compression method. We assess neuron significance based on the absolute magnitude of weights, postulating that layers with greater average weights hold more significance. The sparsity of each layer within the model is established through cross-layer global importance ranking, determining the layer's dynamic compression ratio.}
		\label{fig3}
	\end{figure*}

	Fig. \ref{fig3} illustrates our policy for dynamic gradient compression. Specifically, we assess the importance of each layer based on the magnitude of its weight, $i.e.$,
	\begin{equation}\label{eq3}
		\mu_l^i=\left|w_l^i\right|,
	\end{equation}
	where $w_l^i$ is the $i$-th weight of layer $l$, $|\cdot|$ returns the absolute value of its input. We then conduct a cross-layer ranking across the whole network by the numerical order of $\mu_l^i$. Assuming a global compression ratio $p$, the compression ratio of each layer can be easily determined by sparsifying the weights with the lowest ranks. As a result, each weight $w_l^i$ will be transformed into a sparse representation $\widehat{w}_l^i$ whose elements are defined as:
	
	\begin{equation}\label{eq4}
		\widehat{w}_l^i=\left\{\begin{array}{lc}
			0 & \mu_l^i \text { is among the lowest-rated, } \\
			w_l^i & \text { otherwise. }
		\end{array}\right.
	\end{equation}
	
	Previous work has indicated that more than 99$\%$ of gradients can be safely removed by gradient sparsity with little impact on training accuracy \cite{DGC}. We further consider the importance ranking relationship of different layers of the network, so that the global redundant information can be tracked more accurately, which provides useful information for the subsequent gradient compression.
	
	A straightforward method for quantifying the importance of each layer of the network is to determine per-layer compression rate $p_l$ based on the per-layer sparsity after above operation as:
	\begin{equation}\label{eq5}
		p_l=\frac{\sum_i \delta\left(\widehat{w}_l^i \neq 0\right)}{G_l},
	\end{equation}
	where $G_l$ is the total number of gradients in $l$-th layer, and $\mathcal{\delta(\cdot)}$ is an indicator function, which returns 1 if the input is true, and 0 otherwise. The definition of $p_l$ in Eq.(\ref{eq5}) equalizes the sparsity of the $l$-th layer. 
	To optimize training efficiency, we refrain from executing the aforementioned operation in every iteration. Instead, we determine the frequency of implementing the dynamic compression ratio algorithm based on several factors such as computing resources, efficiency requirements, downstream tasks, and other relevant considerations.
	
	\begin{algorithm}[!h]
		\caption{Algorithm of RS-DGC}
		\label{alg}
		\renewcommand{\algorithmicrequire}{\textbf{Input:}}
		\renewcommand{\algorithmicensure}{\textbf{Output:}}
		
		\begin{algorithmic}[1]
			\REQUIRE The global compression ratio $p$, raw gradient $g$, hyperparameter $\alpha$.
			\ENSURE Sparse gradient $g_s$.
			
			\FOR{each epoch}
			\FOR{$l=1 \rightarrow L$}
			\FOR{$i=1 \rightarrow n_l$}
			\STATE Calculate the weight importance $\mu_l^i$ via Eq.\ref{eq3};
			\ENDFOR
			\ENDFOR
			\STATE Conduct a global ranking of weights by $\mu_l^i$;
			\STATE Obtain the sparse weights by removing the bottom-ranked weights by $p$ via Eq.\ref{eq4};
			
			\FOR{$i=1 \rightarrow L$}
			\STATE Obtain the per-layer compression ratio $p_l$ via Eq.\ref{eq5};
			\ENDFOR
			
			\FOR{each iteration}
			\FOR{$l=1 \rightarrow L$}
			\FOR{$k=1 \rightarrow K_l$}
			\STATE $NSI_p^k\left(x_0, y_0\right)=\alpha \times mean_p^k\left(x_0, y_0\right)+(1-\alpha) \times std_p^k\left(x_0, y_0\right)$
			\ENDFOR
			\STATE Conduct a ranking of $NSI_p^k\left(x_0, y_0\right)$
			\ENDFOR
			\STATE Obtain the sparse gradients $g_s$ by preserving the gradients in neighborhood with the top-ranked $NSI$;
			\ENDFOR
			\ENDFOR
			\RETURN $g_s$
		\end{algorithmic}
	\end{algorithm}

	\subsection{Neighborhood Statistics Indicator (NSI)}
	In order to effectively prioritize gradients for transmission, we introduce NSI as a quantifiable measure to assess the significance of gradients. Most existing gradient compression methods measure the importance of gradients by each individual gradient itself. Differently, we consider a gradient and its neighborhood as a whole.
	
	Generally, a convolution layer transforms an input tensor $\mathcal{X} \in \mathcal{R}^{C \times H_{\text {in }} \times W_{\text {in }}}$ into an output tensor $\mathcal{Y} \in \mathcal{R}^{N \times H_{\text {out }} \times W_{\text {out }}}$ by using the filters $\mathcal{W} \in \mathcal{R}^{N \times C \times K_{h} \times K_{w}}$. Here, $C$ is the number of the input feature maps, $N$ is the number of the filters, and $K_h$ and $K_w$ are the height and width of a filter, respectively. Since weights and gradients have a one-to-one correspondence, the gradient shape of each convolution layer is consistent with the weight, i.e., $g \in \mathcal{R}^{N \times C \times K_{h} \times K_{w}}$. We convert the gradients of each layer into tensors on a two-dimensional plane. Let $g_0$ be a two-dimensional spatial plane coordinate $(x_0,y_0)$. The Chebyshev distance between $g_0$ and $(x,y)$ on the two-dimensional plane is defined as:
	\begin{equation}\label{eq6}
		D_{\text {Chebyshev }}=\max \left(\left|x-x_0\right|,\left|y-y_0\right|\right).
	\end{equation}
	
	The neighboring gradients $G_p (x_0,y_0)$ defined by Chebyshev distance is a set of gradients that are within $p \times p$ patch include $(x_0,y_0)$ itself and is defined as follows:
	\begin{equation}\label{eq7}
		G_p\left(x_0, y_0\right)=\left\{G(x, y)\Big|\left|x-x_0\right|\leq \frac{p}{2},\left| y-y_0\right|\leq \frac{p}{2}\right\},
	\end{equation}
	where $G(x,y)$ is the gradient at position $(x,y)$ in layer $l$. We usually regard the gradient value corresponding to a convolution kernel as a whole. For example, for a $3 \times 3$ convolution kernel, we usually set $p=3$. This is because in DNNs, each neuron is independent, and treating it as a whole can effectively measure the importance of each neuron, rather than treating each gradient value in isolation. The neighborhood defined by Chebyshev distance is shown in Fig. 1, where $p=3$.
	
	We measure the importance of gradients in neighborhood $G_p (x_0,y_0 )$ by calculating the NSI:
	\begin{equation}\label{eq8}
		NSI_p\left(x_0, y_0\right)=\alpha \times mean_p\left(x_0, y_0\right)+(1-\alpha) \times std_p\left(x_0, y_0\right),
	\end{equation}
	where $\alpha$ is the hyperparameter between 0 and 1, and it can be adjusted for different models. $mean_p\left(x_0, y_0\right)$ and $std_p\left(x_0, y_0\right)$ are the mean of absolute value and standard deviation of the gradients in the neighborhood $G_p\left(x_0, y_0\right)$. It is worth noting that we have opted to utilize standard deviation instead of variance in our study. The main reason for this choice is that standard deviation and mean share the same unit. 
	They are defined as follows:
	\begin{equation}\label{eq9}
		mean_p\left(x_0, y_0\right)=\left\|\frac{1}{p^2} \sum_{i=1}^{p^2}\left|g_i\right|\right\|_2,
	\end{equation}
	\begin{equation}\label{eq10}
		std_p\left(x_0, y_0\right)=\left\|\frac{1}{p^2} \sum_{i=1}^{p^2}\left(g_i-\frac{1}{p^2}\sum_{j=1}^{p^2} g_j\right)^2\right\|_2^\frac{1}{2}.
	\end{equation}
	
	Suppose we get the compression ratio corresponding to the $n$-th node and the $l$-th layer of the model $p_n^l$, and then we calculate the number of gradients that need to be preserved:
	\begin{equation}\label{eq11}
		k_n^l=\left\lceil\frac{N \times C \times K_h \times K_w}{p \times p} \times p_n^l\right\rceil.
	\end{equation}
	
	Then, we will preserve all gradient values in the neighborhood corresponding to the Top-$k$ NSI value and transmit the corresponding neighborhood index.
	
	Therefore, our approach can be seen as a variant of Top-$k$ sparsity. It is worth noting that our method differs from Top-$k$ sparsity in two ways: i) By introducing gradient neighborhood, we aggregate all the gradients within the specified neighborhood as a unit to mitigate the contingency caused by individual gradients. This approach effectively minimizes the influence of random noise, resulting in smoother and more stable gradients. ii) Statistic indicators, particularly the standard deviation, assist us in capturing significant yet relatively small gradients, thereby preventing the model from overfitting. Moreover, these small gradients encompass crucial fine-tuning information regarding the model's weight, playing a major role in sustaining optimal model performance.
	
	To solve the problem of gradient optimization direction deviation caused by gradient accumulation directly, Lin \textit{et al.} \cite{DGC} proposed local gradient accumulation using momentum correction:
	\begin{equation}\label{eq12}
		\begin{gathered}
			u_{k, t}=m u_{k, t-1}+g_{k, t}, \\
			v_{k, t}=v_{k, t-1}+u_{k, t}, \\
			w_{t+1}=w_t-\eta \sum_{k=1}^N \operatorname{sparse}\left(v_{k, t}\right),
		\end{gathered}
	\end{equation}
	where the first two terms are the corrected local gradient accumulation, and the accumulation result $v_{k, t}$ is used for the subsequent sparsification and communication. 
	We have also adopted this idea.
	
	\section{Experiments}
	\label{p4}
	\begin{figure*}[t]
		\centering
		\includegraphics[width=2\columnwidth]{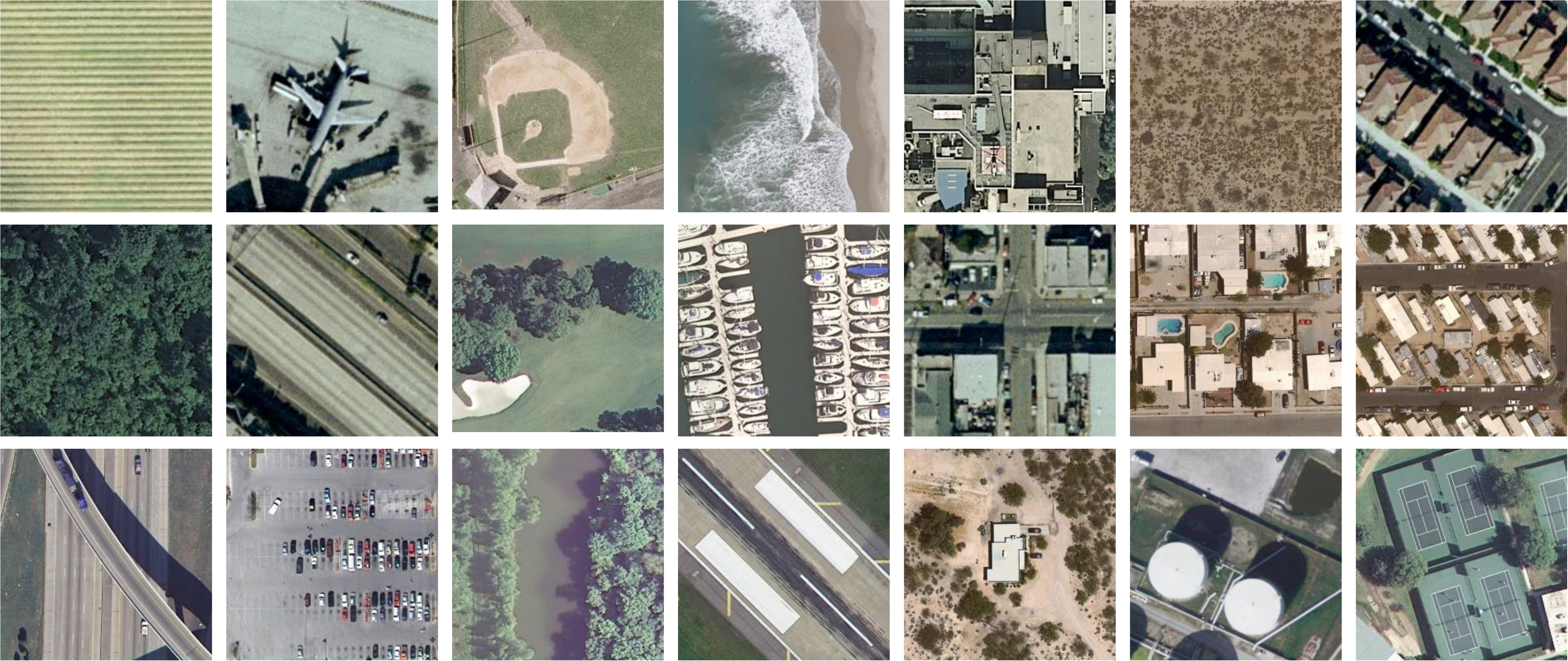} 
		\caption{Example images from the UCML-21 dataset containing 21 land-use scene categories: agricultural, airplane, baseball diamond, beach, buildings, chaparral, dense residential, forest, freeway, golf course, harbor, intersection, medium residential, mobile homepark, overpass, parking lot, river, runway, sparse residential, storage tanks, and tennis court.}
		\label{ucml}
	\end{figure*}

	\begin{figure*}[t]
		\centering
		\includegraphics[width=2\columnwidth]{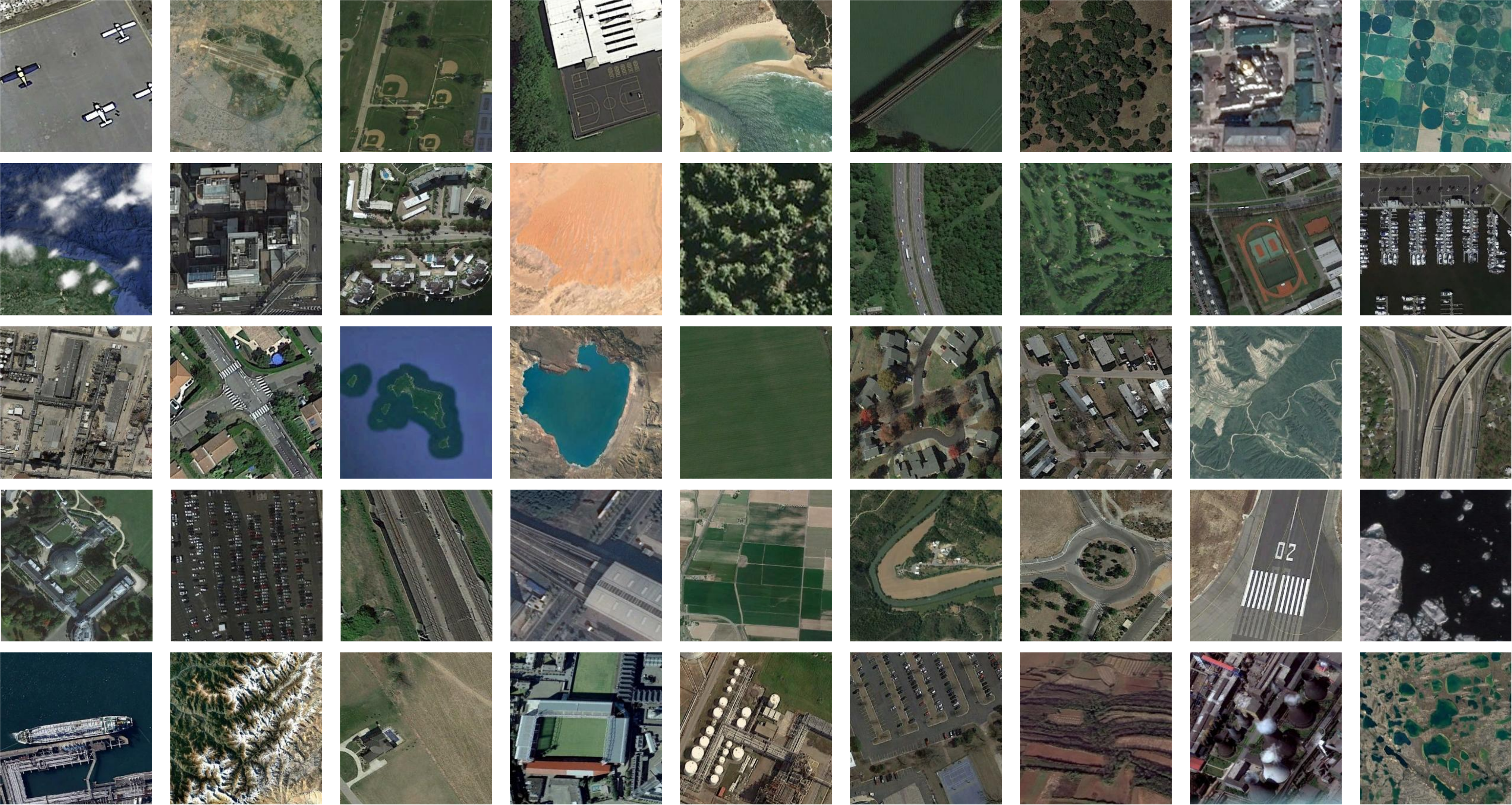} 
		\caption{Example images from the NWPU-45 dataset containing 45 land-use scene categories: airplane, airport, baseball diamond, basketball court, beach, bridge, chaparral, church, circular farmland, cloud, commercial area, dense residential, desert, forest, freeway, golf course, ground track field, harbor, industrial area, intersection, island, lake, meadow, medium residential, mobile home park, mountain, overpass, palace, parking lot, railway, railway station, rectangular farmland, river, roundabout, runway, sea ice, ship, snowberg, sparse residential, stadium, storage tank, tennis court, terrace, thermal power station, and wetland.}
		\label{nwpu}
		\vspace{-0.3cm}
	\end{figure*}
	
	\begin{figure}[t]
		\centering
		\includegraphics[width=0.95\columnwidth]{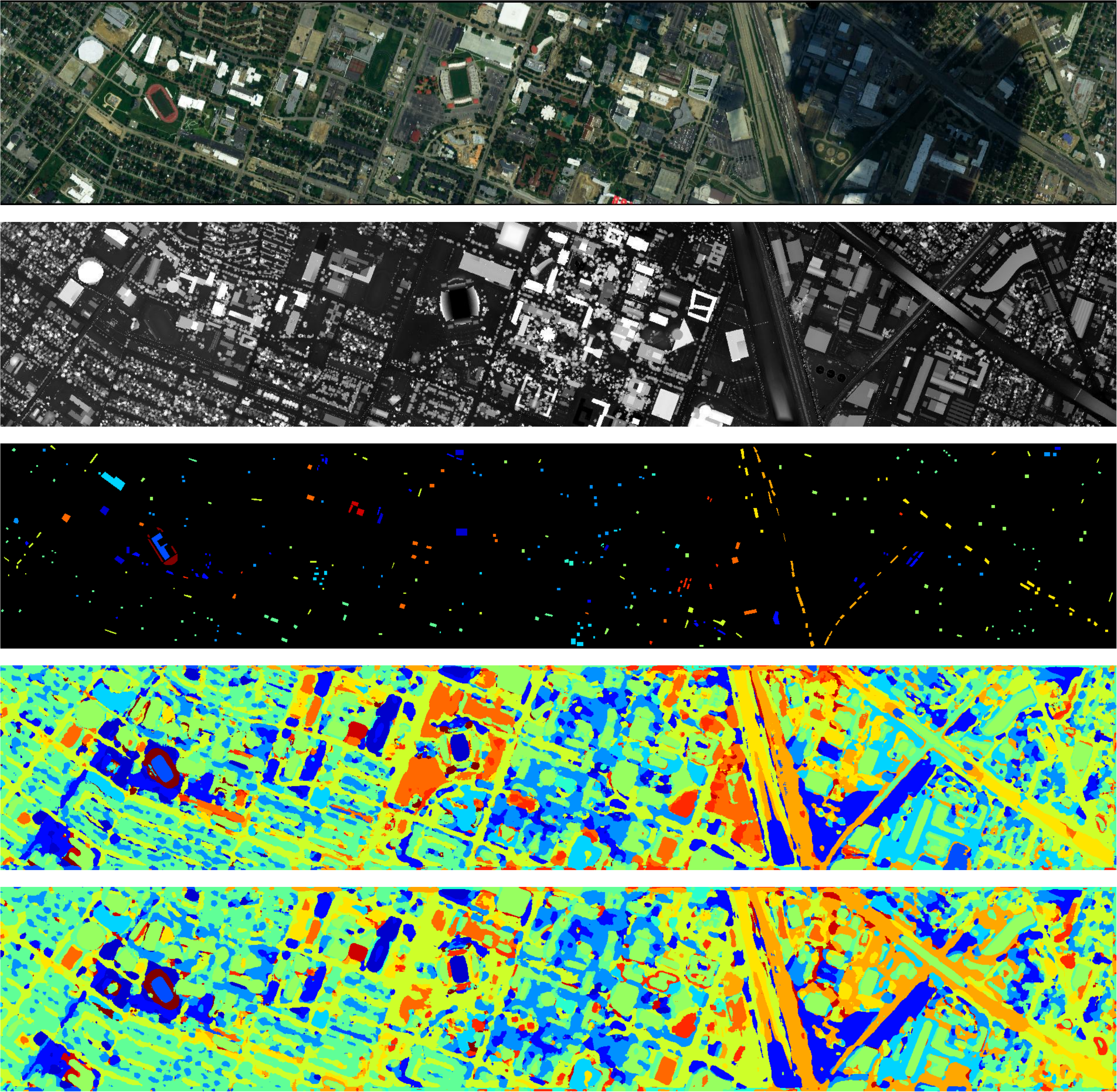} 
		\caption{Scene display and classification results visualization of the Houston2013 dataset. From top to bottom, the HSI of the Houston2013 dataset, the LiDAR image, the labels, the scene classification results using ExViT, and the classification results after RS-DGC.}
		\label{UH}
	\end{figure}
	
	
	In this section, we conduct extensive experiments on two classical downstream tasks in the RS field to evaluate the comprehensive performance of the proposed method. Specifically, it includes RS image classification and multi-modal scene classification. We first present the datasets in Section \ref{4.1}. Then, we introduce the evaluation metrics and parameter setting in Section \ref{4.2}. In Section \ref{4.3}, we conduct ablation experiments. The effectiveness of each module in the proposed method is further verified. In Sections \ref{4.4} and \ref{4.6}, the performance of RS-DGC on the two downstream tasks is reported, respectively. In addition, a series of comparative experiments are carried out to compare with the existing mainstream methods. Finally, in Section \ref{4.7}, we show the actual communication costs reduction of our method.
	
	\subsection{Data Description}
	\label{4.1}
	\subsubsection{UC Merced Land-Use Dataset}
	The UC Merced Land-Use dataset (UCML-21) \cite{ucml} was released in October 2010 by the Computer Vision Laboratory at the University of California, Merced. The dataset images are derived from manually extracting large images from the U.S. Geological Survey (USGS) National Map Urban Area Imagery Collection.  The image size is $256 \times 256$ pixels and the spatial resolution is 1 foot. A total of 21 categories including agricultural area, aircraft, and baseball field are included, and each category has 100 images. Fig. \ref{ucml} shows some sample images randomly selected from UCML-21.
	
	\subsubsection{NWPU-RESISC45 Dataset}
	The NWPU-RESISC45 remote sensing dataset (NWPU-45) \cite{nwpu} is a large-scale public data set released by Northwestern Polytechnical University for the scene classification of RS images. It is extracted from Google Earth (Google Inc.) and covers more than 100 countries and regions in the world. This dataset contains 45 categories of scenes, and the scene samples of each category are displayed as shown in Fig. \ref{nwpu}. Each category contains 700 images, and the size of each image is $256 \times 256$ pixels. Compared with UCML-21, it has the characteristics of large-scale and rich information contained in images.
	
	\subsubsection{The HSI-LiDAR Houston2013 Dataset}
	The HSI-LiDAR Houston2013 dataset is a valuable earth observation dataset that combines hyperspectral and LiDAR data for in-depth studies on land cover and environmental changes. This dataset was created by CosmiQ Works team and uses high-resolution WorldView-2 satellite imagery. Collected by the ITRES CASI-1500 imaging sensor in 2013, an extensive data collection campaign was conducted in the University of Houston campus area in Texas, USA. The hyperspectral component of this dataset comprises 144 bands, covering a wide wavelength range of 0.38$\mu m$ to 1.05$\mu m$, with a spectral interval of 10$nm$. On the other hand, the LiDAR dataset consists of a single-band image. The size of all images in the dataset is $349 \times 1905$ pixels, as shown in Fig. \ref{UH}.
	

	\subsection{Evaluation Metrics and Parameter Setting}
	\label{4.2}
	
	\subsubsection{Evaluation Metrics}
	In RS image classification tasks, the network performance is measured using Top-1 accuracy. 
	As for the multi-modal scene classification task, we employ three indicators: Overall Accuracy (OA), Average Accuracy (AA), and the Kappa coefficient. OA quantifies the proportion of correctly classified test samples out of the total number of test samples. AA represents the average accuracy across all categories. The Kappa coefficient assesses the consistency between the classification map generated by the model and the provided true values.

\begin{table}[htbp]
	\small
	\setlength{\tabcolsep}{1.1mm}
	\renewcommand{\arraystretch}{1.3}
	\centering
	\caption{Ablation Study Results for VGG-19 and ResNet-56 on UCML-21 Dataset}
	\label{t1}
	\begin{tabular}{cccccc}
		\Xhline{1px}
		\multirow{2}{*}[-1.8ex]{\textbf{Models}}  
		& \multicolumn{3}{c}{\textbf{Optimizations}}          & \multicolumn{2}{c}{\textbf{Accuracy}($\boldsymbol{\%}$)}               \\
		\Xcline{2-6}{0.5px}   
		& \textbf{Neighbor}         & \textbf{Statistic}          & \textbf{Dynamic}         & \textbf{Top-1 Acc.} & \thead{\textbf{Top-1} \\ \textbf{Acc.↑} }  \\
		\Xhline{1px}
		\multirow{4}{*}[-0.4ex]{VGG-19}  
		&                  &                    &                 & 87.96      &                   \\
		\Xcline{2-6}{0.5px}
		& $\checkmark$     &                    &                 & 87.98      & 0.02              \\
		\Xcline{2-6}{0.5px}
		& $\checkmark$     & $\checkmark$       &                 & 88.03      & 0.05              \\
		\Xcline{2-6}{0.5px}
		& $\checkmark$     & $\checkmark$       & $\checkmark$    & 88.57      & 0.54              \\
		\Xcline{1-6}{0.5px}
		\multirow{4}{*}[-0.4ex]{ResNet-56} 
		&                  &                    &                 & 90.28      &                   \\
		\Xcline{2-6}{0.5px}
		& $\checkmark$     &                    &                 & 90.31      & 0.03              \\
		\Xcline{2-6}{0.5px}
		& $\checkmark$     & $\checkmark$       &                 & 90.65      & 0.34              \\
		\Xcline{2-6}{0.5px}
		& $\checkmark$     & $\checkmark$       & $\checkmark$    & 91.45     & 0.80              \\
		\Xhline{1px}            
	\end{tabular}
\end{table}
	\subsubsection{Parameter Setting}
	The proposed gradient compression method is implemented with PyTorch \cite{pytorch} on 4 NVIDIA GTX 3090 GPUs and we use NCCL as communication backend. In the training phase, all the networks are optimized by SGD optimizer with a momentum of 0.9. The learning rate is updated by MutilStepLR strategy ($\gamma=0.1$) when using VGGNet and CosineAnnealingLR when using ResNet. 
	For the RS image datasets UCML-21 and NWPU-45, the network is trained for 500 epochs. The initial learning rate is set to 0.1 and the batch size is 16. 
	In the multi-modal scene classification task, we set the batch size to 64 and train for 300 epochs.
	The gradient sparsity is 99.9$\%$ (only 0.1$\%$ is non-zero), and the quantization bit width in QSGD is set to 1 bit.

	\subsection{Ablation Study}
	\label{4.3}
	We conduct ablation experiments using the UCML-21 dataset. Networks used include VGG-19 \cite{VGG} and ResNet-56 \cite{ResNet}. The results are shown in Table \ref{t1}. 
	\subsubsection{Neighborhood information}
	\label{}
	We first verified the importance of neighborhood gradient information. By comparing the first and second rows of VGG-19 and ResNet-56 in Table.\ref{t1}, we can see that neighborhood gradients boost VGG-19 and ResNet-56 by 0.02$\%$ and 0.03$\%$, respectively.
	
	\begin{figure*}[t]
		\centering
		\includegraphics[width=1.8\columnwidth]{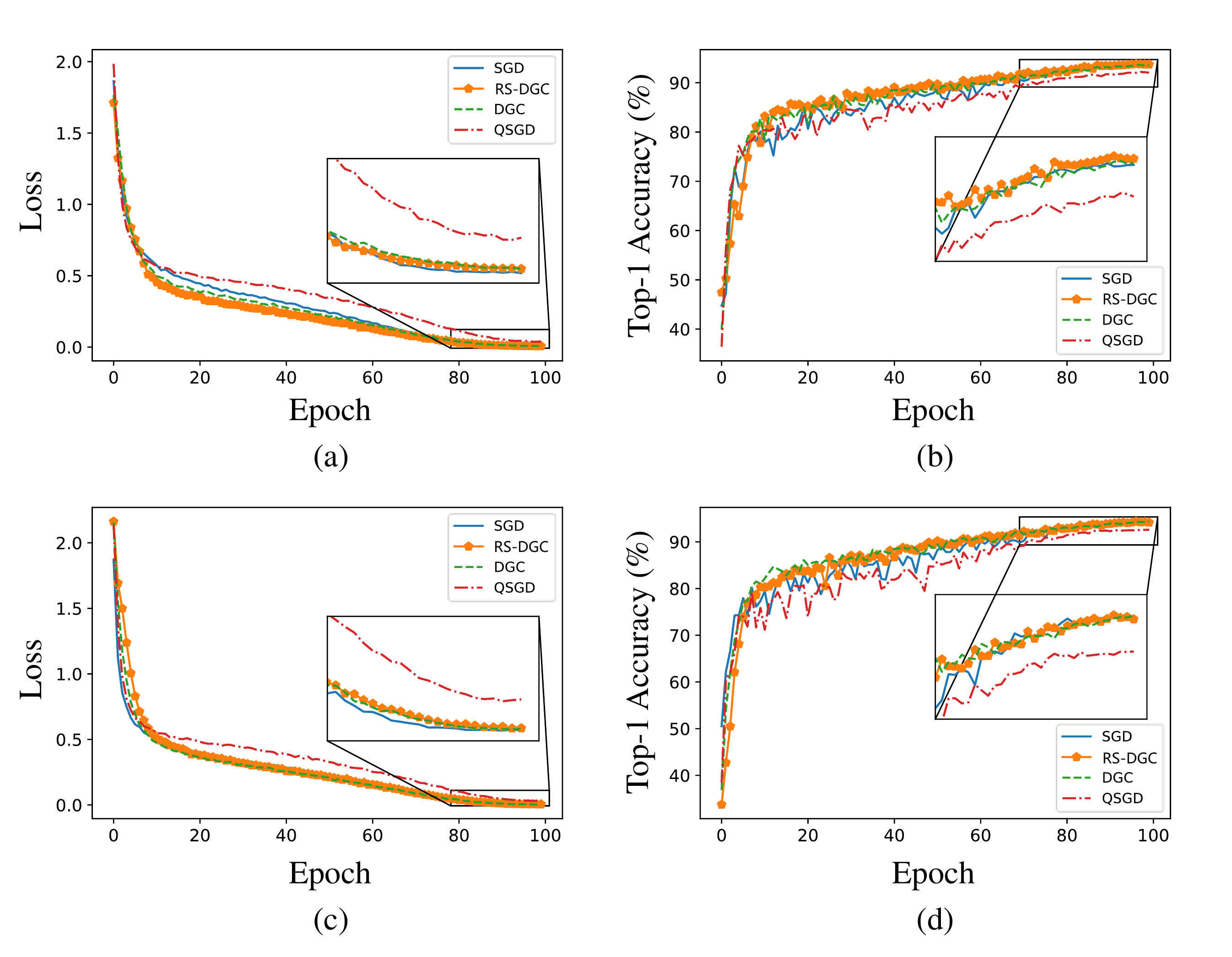} 
		\caption{Learning curves of ResNet-110 and VGG-19 on UCML-21 and NWPU-45 datasets. (a) Training loss of ResNet-110 on UCML-21. (b) Top-1 accuracy of ResNet-110 on UCML-21. (c) Training loss of VGG-19 on NWPU-45. (d) Top-1 accuracy of VGG-19 on NWPU-45.}
		\label{fig5}
	\end{figure*}
	
	\subsubsection{Statistic indicator}
	\label{}
	To verify the effectiveness of statistic indicator, we compared the Top-1 accuracy on UCML-21 datasets with or without standard deviation. 
	As shown in Table \ref{t1}, NSI helps improve VGG-19 and ResNet-56 by 0.05$\%$ and 0.34$\%$.
	
	
	\subsubsection{Dynamic compression}
	\label{}
	Our third experiment combines fixed compression ratio and our dynamic compress strategy on VGG-19 and ResNet-56. The fixed compression ratio $p=0.001$ and we also set the global compression ratio as 0.001 in our dynamic compress strategy. As Table \ref{t1} lists, VGG-19 and ResNet-56 obtain 0.54$\%$ and 0.80$\%$ accuracy improvement over fixed compression ratio respectively, showing the dynamic compress strategy is effective.


	\begin{table}[htbp]
		\small
		\setlength{\tabcolsep}{3mm}
		\renewcommand{\arraystretch}{1.6}
		\centering
		\caption{The Top-1 Accuracy and Gradient Sparsification Ratio on UCML-21 Dataset}
		\label{t2}
		\begin{tabular}{ccccc}
			\Xhline{1px}
			\textbf{Models} & \textbf{Methods} & \multicolumn{2}{c}{\textbf{Top-1 Acc.}($\boldsymbol{\%}$)}  & \thead{\textbf{Sparsification}\\\textbf{Ratio}} \\
			
			\Xhline{1px}
			
			\multirow{4}{*}[-2.5ex]{VGG-19}  
			& SGD  &  92.38  &                &  1$\times$        \\
			\Xcline{2-5}{0.5px}
			& Top-$k$ \cite{top-k}     & 86.01    & -6.37   &  1000$\times$    \\
			\Xcline{2-5}{0.5px}
			& DGC\cite{DGC}     & 86.19    & -6.19    &  1000$\times$    \\
			\Xcline{2-5}{0.5px}
			& QSGD \cite{qsgd}     &  87.27    &  -5.11 & 32$\times$    \\
					\Xcline{2-5}{0.5px}
			& RS-DGC   & \textbf{88.57}   & \textbf{-3.81}   &  1000$\times$    \\
			\Xcline{1-5}{0.5px}
			
			\multirow{4}{*}[-2.5ex]{ResNet-56}  
			& SGD   & 95.24   &        &  1$\times$           \\
			\Xcline{2-5}{0.5px}
			& Top-$k$ \cite{top-k}    & 90.89   & -4.35     & 1000$\times$                   \\
			\Xcline{2-5}{0.5px}
			& DGC \cite{DGC}     & 91.43     & -3.81     & 1000$\times$       \\
			\Xcline{2-5}{0.5px}
			& QSGD \cite{qsgd}    & 90.77     & -4.47  & 32$\times$                     \\
			\Xcline{2-5}{0.5px}
			& RS-DGC   & \textbf{91.45}  & \textbf{-3.79} & 1000$\times$              \\
			\Xcline{1-5}{0.5px}
			
			\multirow{4}{*}[-2.5ex]{ResNet-110}  
			& SGD   & 95.71      &         &  1$\times$               \\
			\Xcline{2-5}{0.5px}
			& Top-$k$ \cite{top-k}     & 93.10   & -2.61  & 1000$\times$                    \\
			\Xcline{2-5}{0.5px}
			& DGC \cite{DGC}    & 93.33  &-2.38   & 1000$\times$            \\
			\Xcline{2-5}{0.5px}
			& QSGD \cite{qsgd}    & 94.79   &-0.92    &  32$\times$                    \\
			\Xcline{2-5}{0.5px}
			& RS-DGC   & \textbf{95.24}   &  \textbf{-0.47}   &  1000$\times$             \\
			\Xhline{1px}
		\end{tabular}
		
	\end{table}

	\begin{table}[]
		\small
		\renewcommand{\arraystretch}{1.6}
		\centering
		\setlength{\tabcolsep}{3.6mm}
		\caption{The Top-1 Accuracy on NWPU-45 Dataset}
		\label{t3}
		\begin{tabular}{cccc}
			\Xhline{1px}
			\textbf{Dataset}                  &  \multicolumn{3}{c}{\textbf{NWPU-45}}  \\
			\Xhline{0.5px}
			
			\textbf{Method}     & \textbf{VGG-19}          & \textbf{ResNet-56 }     &\textbf{ ResNet-110  }      \\
			\Xhline{1px}
			SGD        & 94.06     & 91.08  & 86.29      \\
			\Xhline{0.5px}
			Top-$k$ \cite{top-k}   &  93.88       & 87.11    & 83.17      \\
			\Xhline{0.5px}
			DGC \cite{DGC}      &  94.44       &  \textbf{87.27}       & 83.68         \\
			\Xhline{0.5px}   
			QSGD \cite{qsgd}   &  90.18      & 85.99     &   82.16         \\  
			\Xhline{0.5px}   
			RS-DGC     & \textbf{94.57}     & 87.21      & \textbf{84.69}  \\  
			\Xhline{1px}       
		\end{tabular}
		
	\end{table}

%
	
	\begin{table*}[]
		\small
		\renewcommand{\arraystretch}{1.6}
		\centering
		\setlength{\tabcolsep}{5mm}
		\caption{The OA, AA, and Kappa on the HSI-LiDAR Houston2013 Dataset}
		\label{t5}
		\begin{tabular}{ccccccc}
			\Xhline{1px}
			\textbf{Dataset}   &  \multicolumn{6}{c}{\textbf{The HSI-LiDAR Houston2013 dataset}}  \\
			\Xhline{0.5px}
			\textbf{Network}  &   \multicolumn{3}{c}{\textbf{CCR-Net} \cite{wu2021convolutional}}  &\multicolumn{3}{c}{\textbf{ExViT} \cite{yao2023extended}} \\
			\Xhline{0.5px}
			\textbf{Method}& \textbf{OA}($\boldsymbol{\%}$)   &\textbf{AA}($\boldsymbol{\%}$) &\textbf{Kappa}($\boldsymbol{\%}$) & \textbf{OA}($\boldsymbol{\%}$)   &\textbf{AA}($\boldsymbol{\%}$) &\textbf{Kappa}($\boldsymbol{\%}$)\\
			\Xhline{1px}
			SGD   & 93.15  & 93.16   & 92.56  & 91.40  &  92.60   & 90.66   \\
			\Xhline{0.5px}
			Top-$k$ \cite{top-k} & 88.02  & 87.99   & 87.11   & 82.14    & 82.27   & 81.13  \\
			\Xhline{0.5px}
			DGC \cite{DGC}   & 90.11    & \textbf{90.21}   & 89.53  & 84.96   & 86.67     & 85.72 \\
			\Xhline{0.5px}   
			QSGD \cite{qsgd}   & 88.49   & 88.57  & 87.12    & 81.16       & 81.89    & 80.11 \\  
			\Xhline{0.5px}   
			RS-DGC   & \textbf{90.17}  & 90.19   & \textbf{89.59}  & \textbf{85.99}  & \textbf{86.71}  & \textbf{85.78}  \\  
			\Xhline{1px}       
		\end{tabular}
		
	\end{table*}

	
	\begin{figure}[t]
		\centering
		\includegraphics[width=0.9\columnwidth]{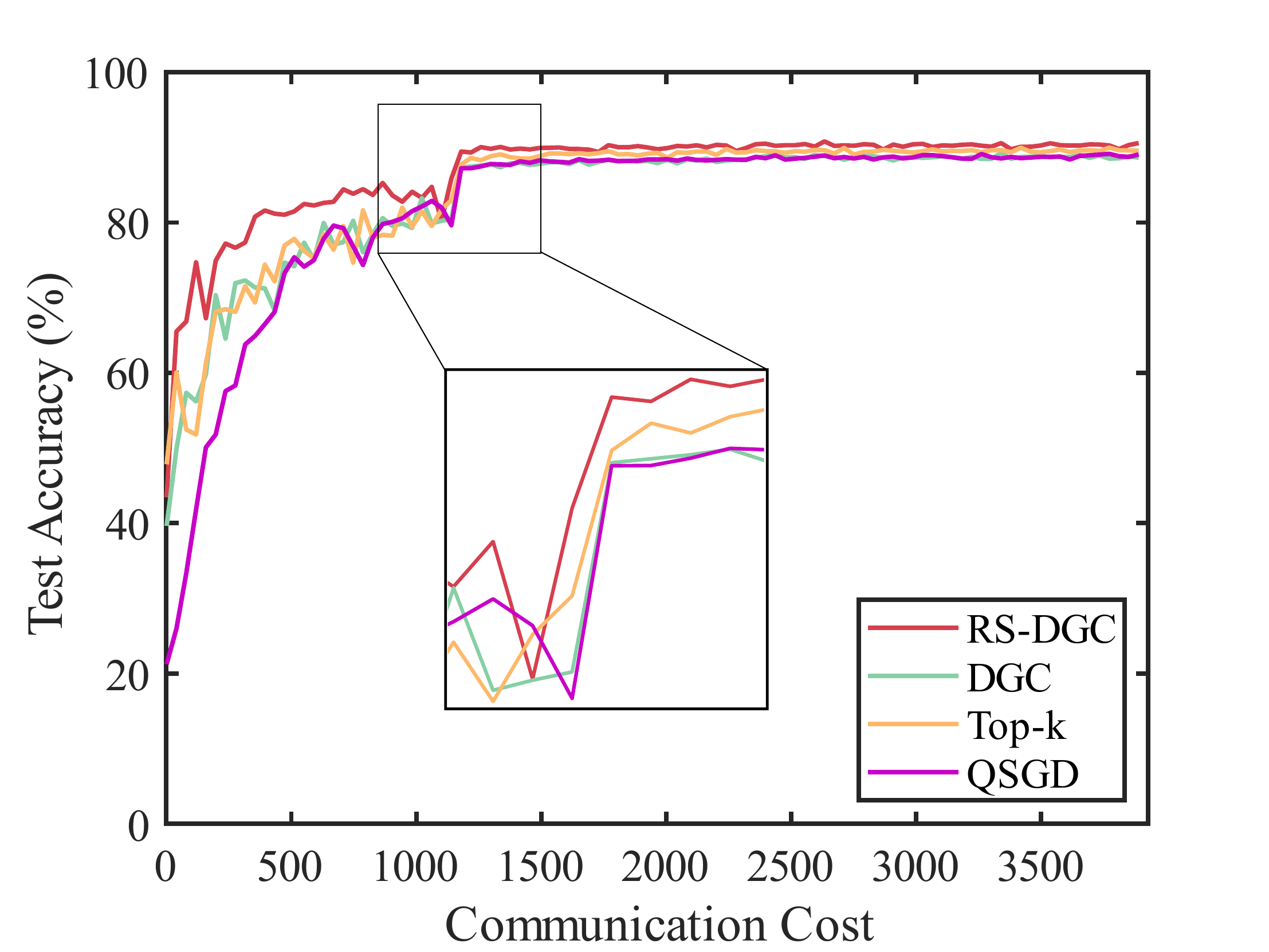} 
		\caption{Comparison of the communication costs for different methods on VGG-19 training on UCML-21 dataset.}
		\label{acc_cost}
	\end{figure}
	
	\subsection{RS Image Classification}
	\label{4.4}
	
	We conduct the RS image classification experiment on UCML-21 and NWPU-45 datasets, with various backbones including VGG-19, ResNet-56 and ResNet-110. We choose the results of distributed SGD optimization as the baseline. 
	
	In the experiment of UCML-21, Table \ref{t2} show the Top-1 accuracy and gradient sparsification ratio compared with other methods. As an improvement of the Top-$k$ method, DGC has indeed improved model performance. However, DGC also has a large accuracy loss on RS image classification. For example, the accuracy loss of VGG-19 is 6.19$\%$, which is not allowed in practice. In contrast, the accuracy loss of our proposed method is only 3.81$\%$ and 3.79$\%$ when using VGG-19 and ResNet-56, respectively. 
	The classification results on NWPU-45 dataset are shown in Table \ref{t3}. Compared to other comparison methods, our method minimizes accuracy loss and even achieves an accuracy improvement of 0.51$\%$ when training the NWPU-45 dataset on the VGG-19 network. One possible explanation we can deduce is that neural networks are overparameterized and their gradient information can be compressed to prevent overfitting of the model for more efficient updates.
	In addition, through the comparison of experimental results between ResNet-56 and ResNet-110 on the two datasets, we observe a gradual decrease in accuracy loss caused by the same gradient compression ratio as the model size increased. This finding indicates that as the model size grows, the presence of redundant information within the model also increases, thereby reducing the impact of compressing the gradient information. Hence, it can be inferred that larger models exhibit a higher tolerance towards gradient compression due to the increased redundancy within them.
	
	Taking VGG-19 and ResNet-110 as examples, we demonstrate the evolution of training loss and Top-1 accuracy in Fig. \ref{fig5}. 
	The learning curve of QSGD is worse than the baseline due to information loss of low-bit representation. It is worth noting that the training loss of RS-DGC decreases faster than other methods. Although the convergence speed of RS-DGC is slower than SGD and DGC at the beginning of training, it converges more rapidly after 10 epochs. As the training process deepens, the stability of our method is better than SGD, and the final training accuracy is better than DGC.
%

	
	\subsection{Multi-Modal Scene Classification}
	\label{4.6}
	To ensure a comprehensive verification of the versatility of our proposed method, we conducted experiments focusing on RS multi-modal scene classification using two advanced multi-modal fusion networks, namely CCR-Net \cite{wu2021convolutional} and ExViT \cite{yao2023extended}, and the results are shown in Table \ref{t5}.
	CCR-Net is a network architecture based on convolutional layers.  After gradient compression, the accuracy loss is small.  Compared with other methods, our proposed method achieves minimal OA and Kappa losses of 2.98$\%$ and 2.97$\%$, respectively. For the transformer-based network architecture ExViT, the accuracy loss of various gradient compression methods is relatively large. This may be due to the fact that the self-attention mechanism layer in ViT involves interactions across all locations. However, when compared with other methods, the method we proposed is still better, with OA, AA and Kappa reaching 85.99$\%$, 86.71$\%$, and 85.78$\%$, respectively. 
	Additionally, we present visualizations of the scene classification results obtained using ExViT. As demonstrated in Fig. \ref{UH}, by comparing the fourth and fifth figures, despite gradient compression, favorable classification outcomes are still attained.

	\subsection{Communication Costs}
	\label{4.7}
	The communication cost and accuracy curves are shown in Fig. \ref{acc_cost}. As we can see, our method has the lowest communication overhead in achieving the same accuracy.
	In addition, we also calculated the actual gradient size that needs to be transmitted in each epoch when training the VGG-16 and VGG-19 networks. As shown in Table \ref{t6}, our method achieves more than 50$\times$ reduction in communication overhead compared to the baseline. At the same time, compared with DGC, our method achieves lower accuracy loss at the same compression ratio.
	
	\begin{table}[]
		\small
		\renewcommand{\arraystretch}{1.6}
		\centering
		\setlength{\tabcolsep}{1.8mm}
		\caption{Gradient Size and Compression Ratio per Epoch for Training VGG-16 and VGG-19 Networks}
		\label{t6}
		\begin{tabular}{ccccc}
			\Xhline{1px}
			\textbf{Network}   &  \multicolumn{2}{c}{\textbf{VGG-16}}   & \multicolumn{2}{c}{\textbf{VGG-19}}\\
			\Xcline{1-5}{0.5px}
			\textbf{Method}&\thead{\textbf{Gradient}\\\textbf{Size(MB)}} & \thead{\textbf{Compression}\\\textbf{Ratio}} & \thead{\textbf{Gradient}\\\textbf{Size(MB)}} & \thead{\textbf{Compression}\\\textbf{Ratio}}  \\
			\Xhline{1px}
			SGD        & 57.19     & 1 $\times$          & 77.45   & 1 $\times$    \\
			\Xhline{0.5px}
			Top-$k$ \cite{top-k}   & 4.23  & 13$\times$    & 5.18   & 15$\times$   \\
			\Xhline{0.5px}
			DGC \cite{DGC}      &  1.19   & 48$\times$        & 1.31     & 59$\times$  \\
			\Xhline{0.5px}     
			\Xhline{0.5px}   
			RS-DGC     & \textbf{1.19}     & \textbf{48}$\boldsymbol{\times}$  & \textbf{1.31}   & \textbf{59}$\boldsymbol{\times}$   \\  
			\Xhline{1px}       
		\end{tabular}
		
	\end{table}


	\section{Conclusion}
	\label{p5}
	In this paper, we propose a novel GC method for RS image interpretation, called RS-DGC, which resolves the contradiction between model performance and high compression ratio with gradient neighborhood and dynamic compression. In particular, we employ NSI to reasonably measure the importance of gradients. Considering the dynamic nature of DNNs training, a dynamic compression strategy is proposed to further enhance the final accuracy. Extensive RS downstream experiments prove the effectiveness and generalization of the proposed RS-DGC.

	\newpage
	\bibliographystyle{IEEEtran}
	\bibliography{reference}
	\newpage
	
	\vfill
	
\end{document}